\documentclass[journal]{IEEEtran}

\usepackage[usenames,dvipsnames]{xcolor}
\usepackage{tcolorbox}
\usepackage{amsmath,epsfig}

\usepackage{amsmath}
\usepackage{amssymb}

\usepackage{array}

\usepackage[english]{babel}
\usepackage{graphicx}
\usepackage{tablefootnote}

\usepackage{caption}
\usepackage{subcaption}
\usepackage{enumitem}
\usepackage{booktabs}

\usepackage[T1]{fontenc}
\usepackage{lmodern}

\usepackage{comment}
\usepackage{color}

\usepackage{amsmath}

\usepackage{array}
\newcolumntype{P}[1]{>{\centering\arraybackslash}p{#1}}
\newcolumntype{M}[1]{>{\centering\arraybackslash}m{#1}}

\makeatletter
\DeclareRobustCommand{\rvdots}{%
  \vbox{
    \baselineskip4\p@\lineskiplimit\z@
    \kern-\p@
    \hbox{.}\hbox{.}\hbox{.}
  }}
\makeatother

\usepackage{graphicx}
\usepackage{booktabs}   
\usepackage[para,online,flushleft]{threeparttable}
\usepackage[]{hyperref}

\usepackage[]{cleveref}   

\usepackage[OT1]{fontenc}    

\usepackage{microtype}  

\usepackage{todonotes}  

\newcolumntype{Y}{>{\RaggedRight\arraybackslash\hspace{0pt}}X}
\newcolumntype{C}{>{\Centering\arraybackslash\hspace{0pt}}X}

\usepackage[acronym,style=long3colheader]{glossaries} 
\usepackage{glossaries}
\usepackage[font=small,labelfont=bf,tableposition=top]{caption}
\usepackage{color, colortbl}

\usepackage{wrapfig}
\usepackage{tabularx}
\usepackage{array}
\usepackage{colortbl}
\tcbuselibrary{skins}

\usepackage{cite}

\DeclareCaptionFormat{myformat}{%
  \begin{varwidth}{\linewidth}%
    \centering
    #1#2#3%
  \end{varwidth}%
}

\newacronym{cnn}   {CNN}   {Convolutional Neural Network}
\newacronym{sgd}   {SGD}   {stochastic gradient descent}
\newacronym{relu}  {RELU}  {Rectified Linear Unit}
\newacronym{gpu}   {GPU}   {Graphics Processing Unit}
\newacronym{klsh}   {KLSH}   {kernelized locality sensitive hashing}
\newacronym{da}  {DA}  {domain adaptation}
\newacronym{rs}  {RS}  {remote sensing}

\DeclareCaptionLabelFormat{andtable}{#1~#2  \&  \tablename~\thetable}

\graphicspath{{./}{images/}}

\title{Accelerated Deep Learning Inference for Large Scale Satellite Imagery in High Performance Computing Environments}

\title{Apache Spark Accelerated Deep Learning Inference for Large Scale Satellite Image Analytics}

\author{\IEEEauthorblockN{Dalton Lunga, Jonathan Gerrand, Hsiuhan Lexie Yang, Christopher Layton and Robert Stewart} \\
\IEEEauthorblockA{National Security Sciences Directorate, \\
Oak Ridge National Laboratory, US}
}

\begin{document}
%
\maketitle
\begin{abstract}
    The shear volumes of data generated from earth observation and remote sensing technologies continue to make major impact; leaping key geospatial applications into the dual data and compute intensive era. As a consequence, this rapid advancement poses new computational and data processing challenges. We implement a novel remote sensing data flow (RESFlow) for advanced machine learning and computing with massive amounts of remotely sensed imagery. The core contribution is partitioning massive amount of data based on the spectral and semantic characteristics for distributed imagery analysis. RESFlow takes advantage of both a unified analytics engine for large-scale data processing and the availability of modern computing hardware to harness the acceleration of deep learning inference on expansive remote sensing imagery. The framework incorporates a strategy to optimize resource utilization across multiple executors assigned to a single worker. We showcase its deployment across computationally and data-intensive on pixel-level labeling workloads. The pipeline invokes deep learning inference at three stages; during deep feature extraction, deep metric mapping, and deep semantic segmentation. The tasks impose compute intensive and GPU resource sharing challenges motivating for a parallelized pipeline for all execution steps. By taking advantage of Apache Spark, Nvidia DGX1, and DGX2 computing platforms, we demonstrate unprecedented compute speed-ups for deep learning inference on pixel labeling workloads; processing 21,028~Terrabytes of imagery data and delivering an output maps at area rate of 5.245sq.km/sec, amounting to 453,168 sq.km/day - reducing a 28 day workload to 21~hours. 
\end{abstract}



%

\section{Introduction}
Earth observation and remote-sensing are both fields that have undergone a renaissance recently, making major impacts in key geospatial applications including land cover mapping, infrastructure mapping, damage assessment, and population distribution studies~\cite{DBLP:journals/corr/abs-1805-08946,Gueguen_2015_CVPR,Hamid_ISPRS2014,Bhaduri2007}.  Multiple factors are contributing to this change, including significant improvements and rapid deployment of satellite technologies that are enabling the acquisition of vast volumes of high resolution imagery at high velocity rates. As such, remote sensing applications have leaped into a data and compute intensive era presenting challenges and opportunities for new advanced machine learning and computer vision workflows. Examples of such applications include providing possibilities to study sustainability outcomes at scale~\cite{Oshri2018}, and identifying urban environments over large contexts using abundant satellite imagery and breakthroughs in deep learning based image classification~\cite{Albert2017}. To achieve greater impact with machine learning on data and compute intense workloads, new advanced workflows are required for efficient utilization of high performance computing resources.

Modern advances in computing hardware are enabling new opportunities for the manner in which large volumes of imagery data are processed. Over the past decade, Hadoop has emerged as an early experimental testbed for several big data applications due to its excellent large-scale data-handling capability, high fault tolerance, reliability, and low cost of operation~\cite{apache_hadoop}. Hadoop provides distributed data storage and analysis solutions which previously have been exploited for implementing large scale mean-shift based image segmentation algorithms~\cite{doi:10.1117/12.2283032}. In~\cite{7436925}, an optimization effort on the Hadoop file storage system was studied to elicit better performance for large scale computing with image data. The authors of ~\cite{10.1007/978-3-642-16515-3_21} study a Hadoop and MapReduce~\cite{Dean2010MapReduceAF} based implementation of the parallel K-means algorithm to reduce the computational time taken for executing  parallel data clustering on a large number of satellite images. Pursuing content mining on digital images, the authors in~\cite{Cao2016} introduced an approach for large-scale scene retrieval on massive image databases.  

While MapReduce enables large-scale distributed computing for imagery when used in conjunction with Hadoop, a limitation is seen in its heavy usage of disk input-output (I/O) operations and network resources to store intermediate steps during processing. Deterred by this computational cost, Huang et al.~\cite{7464827} study Apache Spark~\cite{Zaharia:2010:SCC:1863103.1863113} to take advantage of its resilient distributed datasets (RDDs)~\cite{matei_rdd}. Spark has been shown to accelerate several other remote sensing imagery workloads. For example, in applications where trans-regional remote sensing images are key, the frequent data I/O requirements for mosaicking were shown to benefit from a parallel algorithm implemented with Spark~\cite{Zaharia:2010:SCC:1863103.1863113,8025338}. The work of Sun et al.~\cite{Sun2015} also demonstrates this performance increase, wherein the authors implement an iterative singular value decomposition algorithm to process massive amounts of remote sensing data. In concurrence, high performance computing environments are enabling targeted computing with extremely large earth observation data and the sharing of data in parallel across hundreds of nodes\cite{Pittman:2018,Mathuriya:2018}. Taking advantage of the high processing power, large memory capacity, and Infiniband-enabled interconnects between nodes in Summit, \cite{Kurth:2018} proposed an exascale ready workflow and software stack for extracting signals for extreme weather patterns using variants of deep neural networks. The work scaled up to 27'360 V100 GPUs and sustained throughput of 325.8 PF/s and a parallel efficiency of 90.7\% in single precision.
Another growing technology, even though focused on horizontal scaling resources is cloud computing. The technology is suited for targeting embarrassingly parallel tasks. Leveraging deep feature learning and parallel computing power, in \cite{Sun2019} an efficient and scalable framework for pan-sharpening of remotely sensed big data in cloud computing environments was proposed. The framework characterizes a remote sensing application as a directed acyclic graph for cloud computing with Apache Spark~\cite{Zaharia:2010:SCC:1863103.1863113} and TensorflowonSpark~\cite{yang2017open}. Several copies of data set are distributed via RDDs across virtual computing nodes to perform achieve distributed computing at scale.

Herein, we present RESFlow, an end-to-end workflow to address three challenges in large-scale machine learning for geospatial applications: (1) data complexity, (2) computational complexity, and (3) human labor requirements for collection of ground-truth data. 
\emph{On the data intensive challenge:} The shear volumes of remote sensing imagery are increasingly becoming heterogeneous and challenging the efficacy of current machine learning workflows. With imagery data acquired as signals from varying system configurations and environmental conditions, efforts to analyze such diversity at scale are immediately thwarted by the lack of generalizable workflows. RESFlow seeks to stratify imagery into homogeneous distributions from which levels of diversity are contained to inform the reuse of models for inference tasks on imagery partitions with similar characteristics but  originating from mixed geographies.

\emph{On the compute intensive challenge:} Pixel labeling algorithms are compute intensive, necessitating efficient use of high processing computing resources on more than one computing node, either for feature extractor training or model inferencing. To address computational aspect of the problem, we propose to formulate the application as constituting a set of sub-tasks with interdependence but which are each executable in an embarrassingly parallel fashion. Given that sub-tasks have dependency on each other, this imposes an order of precedence on their execution, creating a task scheduling problem and resource contention that we handle via a novel GPU ticketing system. 

\emph{On the labor intensive challenge:} Current techniques from machine learning, especially deep learning, continue to demonstrate near-human performance. However, such methods are heavily dependent on large amounts of annotated data sets. When small amounts of high-quality labeled data are available, models tend to achieve poor generalization capability. The endeavor to obtain high quality training data can be tedious, especially for pixel labeling, and is characterized by multiple attributes that include; availability of domain experts, stratification of data into diverse representative samples, mitigation of human sampling bias, and accurate labeling of data samples. By seeking automated mapping  into homogeneous partitions, our goal is to create data buckets from which to sample representative images with similar characteristics, mitigating the need to stratify large and diverse training data.

This paper seeks to address the above mentioned challenges from a generic perspective  amenable for use in other large scale object mapping applications using remote sensing imagery. The technical contributions of this paper are: {\em (1)} We propose a novel and efficient single pipeline with multiple deep neural networks for multi-pass distributed image data analysis performed in an embarrassingly parallel fashion. {\em(2)} We present an unprecedented homogeneous partitioning of massive amount of data based on the semantics and spectral characteristics for parallel and distributed imagery analysis.
{\em(3)} We leverage a deep metric space to create unique binary codes for efficient indexing 10s of models and 1000s of image patches for distributed pixel labeling. {\em(4)} We take advantage of Apache Spark to provide, for a single large image scene, a fast parallel inference functionality achieving tremendous speed-up with area pixel labeling rate of 5.245sq.km/sec, amounting to 453,168 sq.km/day - reducing a 28 day workload to 21~hours. {\em(5)} We present a containerized workflow for Apache Spark operations coordinated with GPUs for deep learning inference best practices, e.g. efficient GPU usage and ticketing across multiple workers, for large deep learning workloads deployed on GPU clusters.

Although presented for a pixel labeling task on satellite imagery, the workflow can easily be deployed for bio-medical and climate image based applications.

\section{Satellite Image Analytics with Deep Neural Networks}\label{sec:sequential-workflow}
Given the prevalent nature of high resolution remote sensing instruments, it is now conceivable to pursue computer vision methods for large scale object segmentation. Very-high-resolution remote sensing imagery, which now supports ground spatial resolutions of less than $50$cm, is enabling new capability to exploit subtle and yet expressive spatial features for fitting highly complex objective functions for structured predictions with computer vision and machine learning methods. Deep convolutional neural networks have become the dominant machine learning technique for visual recognition, achieving state-of-the-art results on a number of problems that seek dense semantic labelling of image pixels. Early attempts on this problem include work in~\cite{ DBLP:journals/corr/ChenPKMY14}  where an atrous method to expand the support of filters and reduce the down-sampling for input feature maps to achieve dense labeling was used. In~\cite{ronneberger2015u}, an efficient and precise biomedical image segmentation convolutional neural network (U-net) was proposed. Improving on the architectural design to reconstruct the original input resolution, \cite{Badrinarayanan2017} proposed a semantic pixel-wise segmentation method using a fully-convolutional neural network (Seg-Net), which uses decoder-deconvolutional layers to map the low-resolution encoder feature maps to the full input resolution feature maps.
The use of deep convolutional neural networks extends to other applications including big data mining for search and retrieval tasks. Designed to seek expressive spatial and visual content representational features, a deep hashing framework demonstrated capability for large-scale image retrieval in \cite{8067633Li}. In another image retrieval task, pre-trained networks were used in order to extract intermediate image representation as input for metric and hash-code learning\cite{8518381Roy}.

In general, to perform complex tasks at the level of humans, deep learning methods heavily depend upon the availability of enormous amounts of high quality annotated data. Despite the fact that remote sensing instruments are acquiring data in substantial volumes and the robust computing power needed to efficiently process it is available; such massive data sets are not simple to annotate. The process of gathering labeled training data is mired by inconsistencies, poor selection of representative samples and the annotation is often prohibitively expensive. It is labor intensive requiring a huge number of worker hours, making it challenging to train a single high performing deep network model for use on wide area geographical coverage. We therefore feel it is appropriate to seek automated workflows which support representative training data selections e.g. avoid under-representation, enable localized model for capturing homogeneous data distributions or fit models on diverse yet equally sampled image characteristics.

\begin{figure}
    \centering
    \includegraphics[scale=.8]{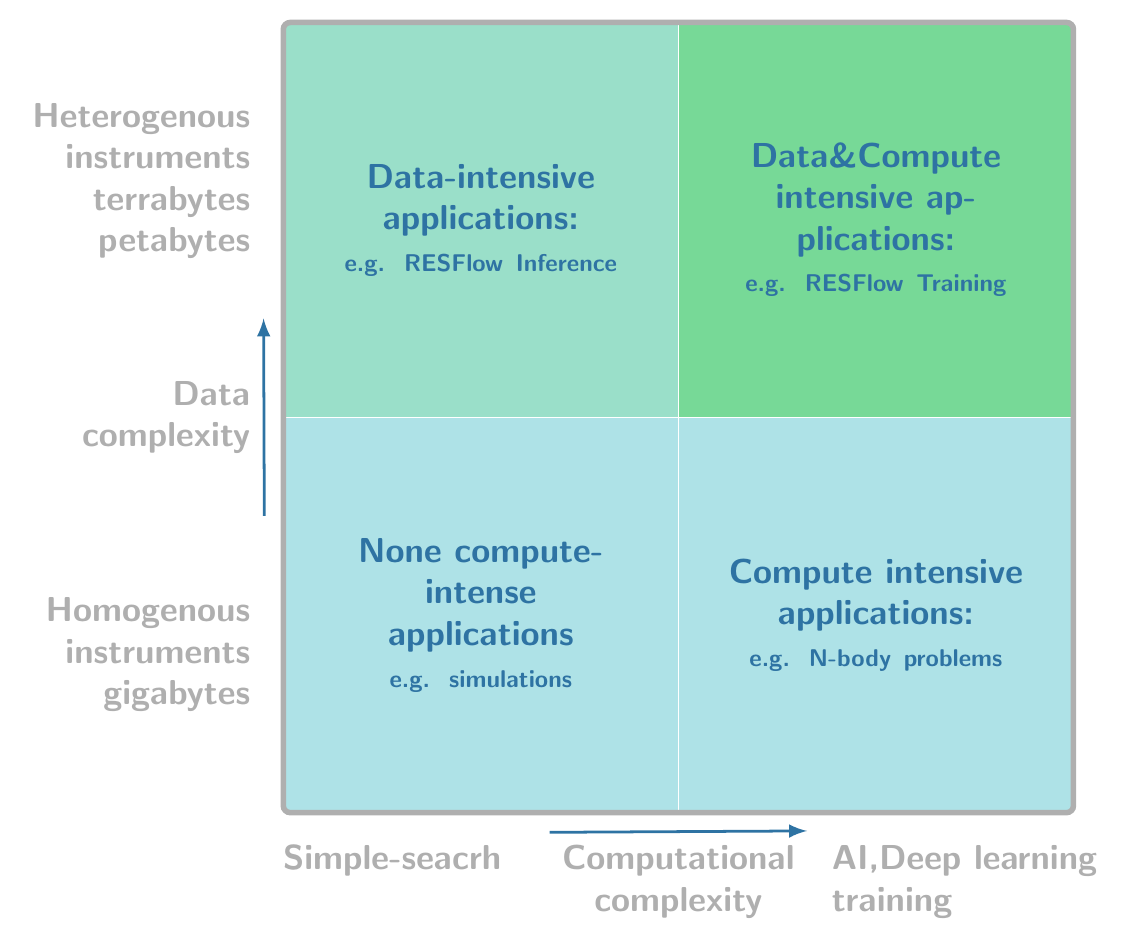}
    \caption{Data, compute and labor intensity paradigm in remote sensing applications. (Note: labor is illustrated with color intensity {\em limegreen} denotes complexity in labor demand.}
    \label{fig:data-compute-intense}
\end{figure}

Moreover, with a growing demand for geospatial applications to deliver imagery products on data that scales over $10$s of Terrabytes~(TB) of high-resolution outputs per given geographic region, current applications are gradually becoming immense in terms of  both data and compute requirements. To be specific, typical workloads for semantic labeling often entail processing imagery acquired across an average country of land-area size $783,562$sq.km. The accompanying imagery coverage would span $3\times783,562$sq.km  to account for scene overlapping and lack of cloud free data. Compare this volume of data against the  largest known computer vision dataset ImageNet\cite{imagenet_cvpr09}. ImageNet has a total of $14,197,122$ images, each at $224\times224$ pixels thus $50,176$ pixels per image, totaling $712,354,793,472$ pixels. When sampled at ground-sampling distance of $50$cm, a mosaicking of all imageNet totals $356,177$sq.km, slightly less the  size of Montana, US. In contrast, for an average country size, at $50$cm ground sampling distance each RGB image scene spans about $40,000\times35,000$ pixels and carries $\approx 13$~GB of data for a total of $3000$ scenes (covering equivalent of $3\times783,562$sq.km ($7\times$ImageNet) land-area and totalling $\approx39$TB of data). Using current serial processing pipelines a single image scene takes $35$-minutes to process a pixel-labeling task on a single computing node with one $16$GB GPU card. Considering the demands to process multiple country scale products, it is imperative that object segmentation and semantic labeling tasks are deployed through parallel and distributed inferencing pipelines to reduce such computational complexity. With this motivation we identify advanced remote sensing dataflows (including \emph{RESFlow}) to be located in the top two quadrants of Figure~\ref{fig:data-compute-intense} and continue to develop core computational modules that can match the demands of such applications.

\section {Proposed RESFlow Framework}\label{sec:resflow} 
\begin{figure}
    \centering
    \includegraphics[scale=0.7]{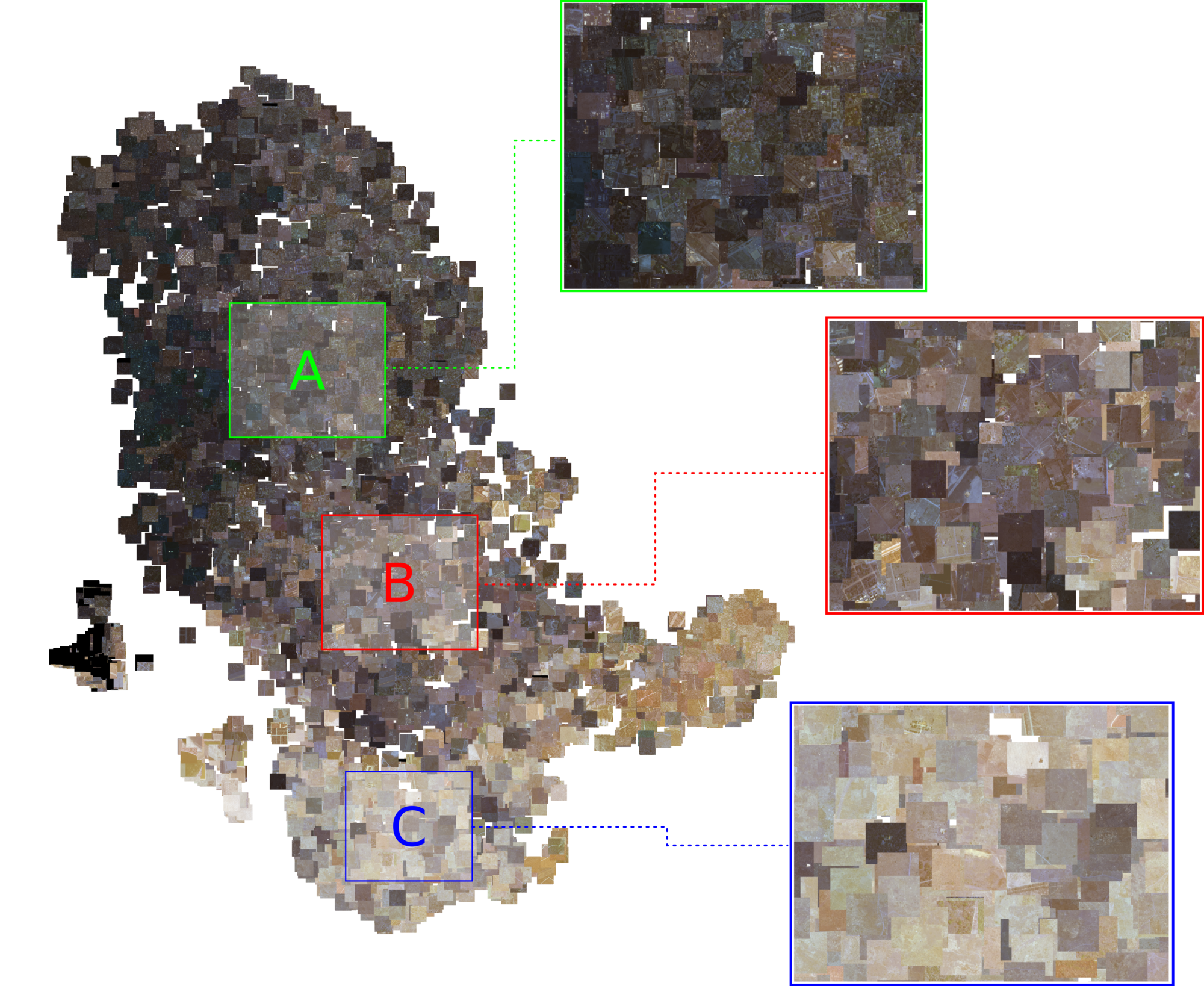}
    \caption{A visual representation of a continuum embedding space for an image scene. Groupings A, B and C illustrate sets of image patches that inform homogeneous data partitions in the workflow.}
    \label{fig:embedding}
\end{figure}

The RESFlow architecture seeks to present itself as an intelligent big data engine where end-to-end inference tasks are efficiently executed while exploiting the geometry of the data and being agnostic to sensor variations as well as geographic constraints. To this end, it is formulated to contain several integrated computational modules and algorithms; providing a common data pipeline which is shared across multiple inference tasks and geospatial applications. At the core of RESFlow is the concept of data distribution partitioning which is performed via efficient geometric based clustering and metric learning. Both techniques play key roles in the overall mapping of data to a partitioned image gallery for indexing - a strategy often lacking in traditional learning workflows. Here we believe that data partitioning is key to mitigating bias in training data collection and labelling (even though not in scope of the current study). Figure~\ref{fig:embedding} illustrates the partitioning of an embedding representation for image patches extracted from a single large satellite image scene. From this result, the concept of data partitioning can be observed as the grouping and extraction of homogeneous image spaces for further exploitation during inference of the large image scene.

As noted above, computer vision and machine learning algorithms are proving superior in providing automated means to describe the distinctive nature of objects in remotely-sensed image data \cite{DBLP:journals/corr/abs-1805-08946,Oshri2018,Albert2017, PRAKASH201892}. However, the deployment of such algorithms remains a significant challenge when considered on large geographic areas covered by hundreds of thousands of images \cite{PRAKASH201892}. 

As is tradition with data/compute intense applications, success depends upon scarce and expensive hardware resources. It is therefore not surprising that use of hybrid CPU/GPU technology stacks is emerging as the means to address such deployment challenges. For example, deep learning functionality for analysis can be developed as user defined functions (UDFs) and used within Apache Spark clusters for inference deployment on GPU and CPU servers in a resourceful manner. Motivated by such potential we combine salient features from the  deep learning frameworks (e.g. TensorFlow and PyTorch) and big-data capabilities from Apache Spark, to implement accelerated and parallelized inference modules for use on both CPU and GPU servers.

\begin{figure*}
  \makebox[\textwidth]{\includegraphics[scale=0.23]{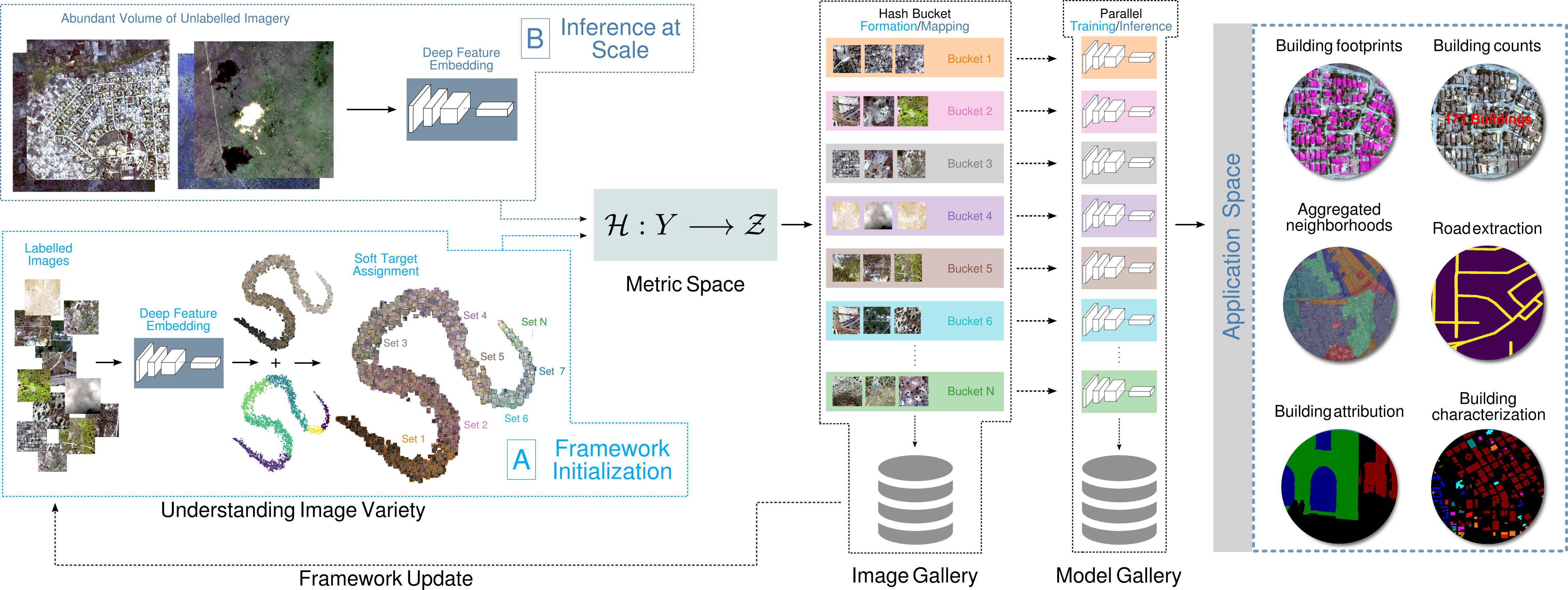}}
    \caption{Overview of the RESFlow framework.}
    \label{fig:resflow-overview}
\end{figure*}
The central means to achieving such capabilities is the idea that both remotely-sensed image data, and deep learning models, can be mapped to and paired within local regions in which the extreme diversity induced by sensor characteristics and scene content is constrained. As depicted in Figure~\ref{fig:resflow-overview}, this procedure is initially facilitated using a learned functional mapping which partitions high dimensional data embeddings into several buckets of similar semantic and spectral content and stored within an image gallery. The bucket partitions then provide a basis to train and update associated indexable models, stored within a model gallery, which can be tailored for specific inference tasks as defined by the application domain.

We briefly describe the \emph{RESFlow} architecture several of its modular components in the following sections.

\subsection{Clustering and embedding module} The first step to enable remote sensing imagery partitioning is utilizing the Clustering and Embedding module (CEM) in \emph{RESFlow} 
The responsibilities of the CEM are two-fold. First, as the initial step within the coalescing of imagery to appropriate partitions of the Image Gallery, the module maps each input image as a datapoint using a learned feature extractor to an intermediate representation in which other datapoints characterized by similar acquisition conditions, spectral and semantic content share a close proximity. This intermediate mapping, or network embedding, is important as it provides a basis for an appropriate metric space to be learned in a data-driven manner. Second, during \emph{RESFlow's} initialization, the CEM is used to assign labels via clustering as a means to assist learning of the metric space projection function.

\subsection{Image-Bucket Assignment}
After clustering the partitioned images, we seek to construct buckets that uniquely represent those clustered images. The Unique binary representation generated for each image plays a significant role in two ways: (1) they provide compact binary bitstrings that preserve semantically similar content for the partitioned image chips and (2) the binary bitstrings provide an efficient image-model indexing mechanism during large scale inference. This dual benefit is achieved by learning a hashing metric space whose properties include 1) generating a unique hash-map associated with each distinct image chip, 2) providing a compact representation which is smaller than the original input dimensions, and 3) a metric space from which the distance property for binary bitstrings can be used to relate image scenes whose geographies and image characteristics are also similar. The resulting binary bitstrings represent a desirable format with which to both efficiently index pre-trained models and the respective buckets  proving an intuitive gathering space for localized training data. 

\begin{figure*}
    \includegraphics[width=\textwidth,height=.6\textwidth]{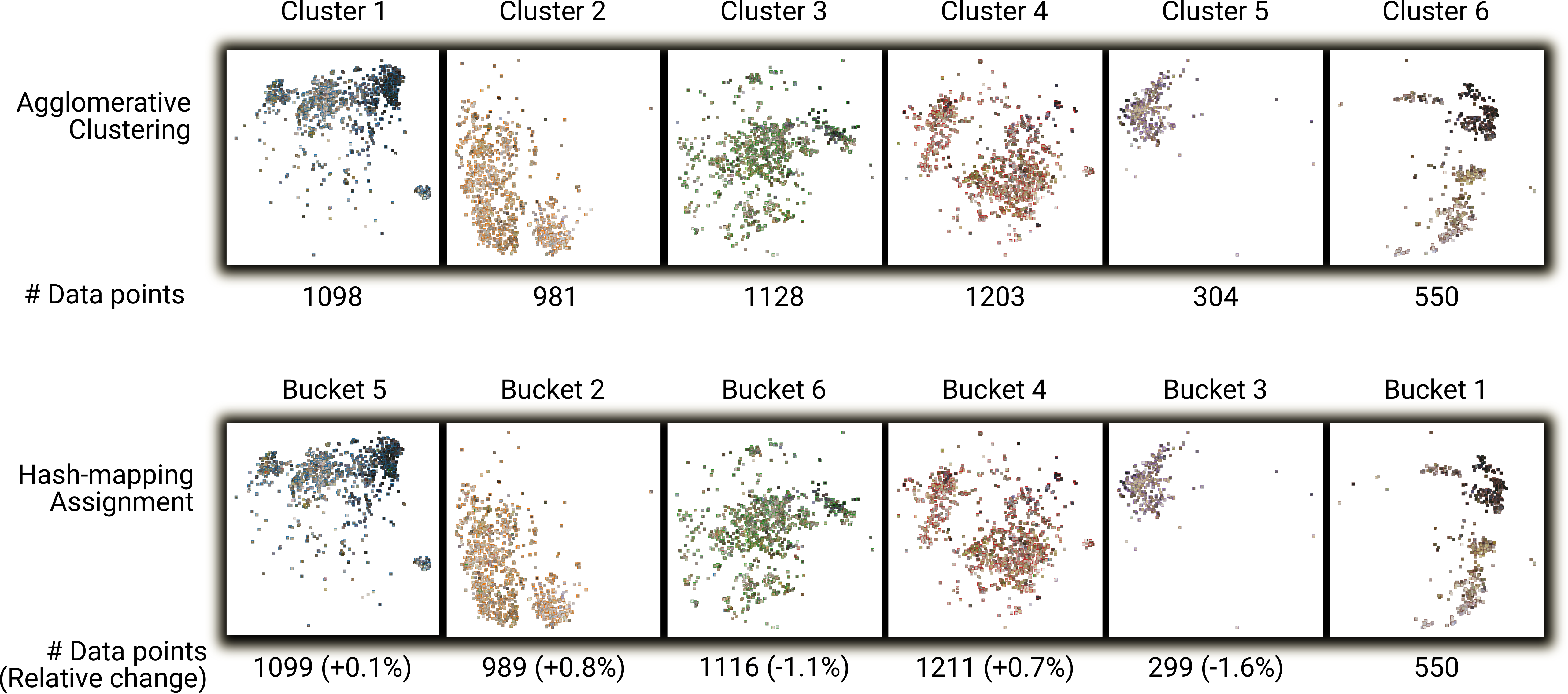}
    \caption{Comparison between buckets initialized via agglomerative clustering and hash-mapping.  Each cluster on the top \textsf{row} is color-coded to denote image patches belonging to one cluster. The bottom \textsf{row} shows a reconstruction of clusters mapped into hash-buckets. Hash-buckets are reconstructed with a validation \textsf{mAP} score of $98.3\%$ with fewer image patches on the bottom row noted to have been placed in different buckets than their cluster of origin.}
    \label{fig:bucket-assignments}
\end{figure*}

The assignment of images to buckets is achieved by first computing the centroid binary bitstring for each bucket. Each bucket centroid offers a unique binary bitstring that's reused to identify each bucket model. Secondly the centroid gets reused within a hamming space to identify which bucket each image chip be assigned to. Figure~\ref{fig:bucket-assignments} illustrates the main concept upon which {\em RESFlow} operates with hash-mapped buckets created via clustering. Clustering provides the key soft-labels needed to learn the semantic structure of the large satellite imagery archive. As shown, the reconstruction of the similar content structure, without emphasizing cluster-bucket correspondence, is carried out by the hash-mapping function. Furthermore, the distinct buckets become completely indexable within the image gallery shown in Figure~\ref{fig:resflow_overview}. 

To illustrate the hash-mapping module we observe the ability of a convolutional network based model to reconstruct the CEM generated embedding space while varying the number of initial clusters. Clusters for generating the soft-labels are chosen based on observing the number of clusters with an average smallest variance per cluster while varying the number of clusters. The hash-mapping network is evaluated using the mean  Average Precision (mAP) metric 
which assesses the average value of the maximum precision for different recall levels while reconstructing the structure of initial clusters. Using a color-coding scheme, Figure~\ref{fig:bucket-assignments} shows the relative changes between the agglomerative based clusters and the convolutional neural network based hash-mapped buckets.


\subsection{Image gallery} 
Following their creation via the metric space projection function, the binary bitstrings generated from input imagery are partitioned by similarity relative to the distinct buckets shown in Figure~\ref{fig:bucket-assignments}. These partitions form a powerful abstraction, independently characterizing homogeneous image acquisition characteristics and spectral content, from which both training and inference data can be sourced for a related Model Gallery network.  Further associated with each binary bitstring entry within this gallery is additional meta-data detailing attributes such as geo-information, acquisition conditions, storage location and image sub-coordinates. Combined with the efficient binary representation of each datapoint for rapid image search capability, the inclusion of these attributes enable rich insights to be gained through exploratory inspection of the Gallery content. These insights could not only benefit performance evaluation of models but also help explain complex data characteristics and their bias.

\subsection{Model gallery}
Once populated, the Image Gallery presents a collection of homogeneous partitions. As previously motivated, this collection, with constrained diversity, is a desirable property when sampling for both training and inference data with a given deep learning algorithm. The Model Gallery will provide a direct mapping between a bucket partition within the Image Gallery and a paired and trained network model which is either fine-tuned during training, or applied during inference. Following this procedure can prove as a highly efficient alternative to standard model training and inference practices, as the localized constraints placed on the new data which each gallery model sees mitigates the need for continual retraining.

\subsection{Accelerated inference}
\begin{figure}
    \centering
    \includegraphics[width=\linewidth]{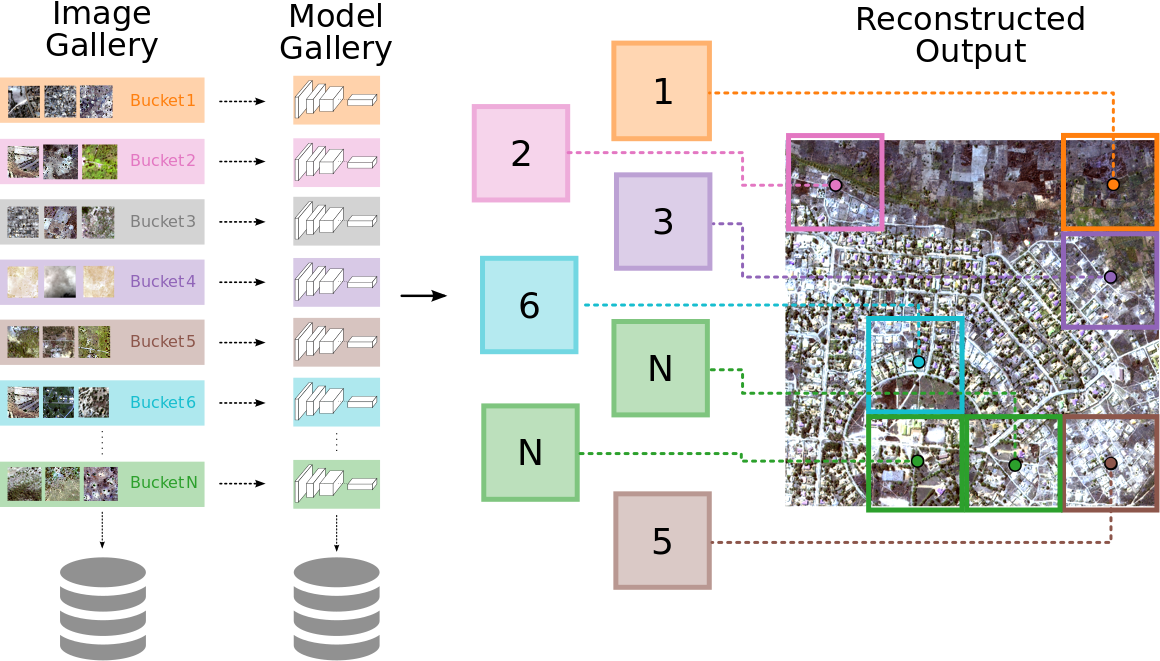}
    \caption{\emph{RESFlow} tile inference and reconstruction illustration. Colors denote inference deployment of different gallery models as assigned by the image-bucket module. The tile partitioning strategy injects a key property that allows for spatially non-contiguous image tiles to be processed by a single bucket model - enabling consistent inference over wide geographic conditions.}
    \label{fig:scene-reconstruction}
\end{figure}

To this point, each of the discussed modules are seen to play a role in overcoming the extreme variance characteristic within imagery with the core concept of partitioning remote sensing imagery. However, the issue of analyzing this data at sufficient scale to meet the demands of global-size applications remains a prohibitive concern. In addressing this problem, it is further observed that each of the modules within \emph{RESFlow} utilize elements of deep learning to perform different tasks, presenting a computationally heavy workload that requires the same GPU hardware resources to be reused in a single inference run. However, \emph{RESFlow}'s building blocks and their functionality are amenable to massive parallelization across the partitioned image space. In recognition of this condition, we exploit Apache Spark as a fabric for distributing and coordinating the framework's computations at scale. Learning from libraries such as TensorflowOnSpark~\cite{yang2017open} and Tensorframes~\cite{hunter2016tensorframes}, we leverage Spark's big data capability to ingest large quantities of input data and process these using deep learning frameworks complimentary to Spark in a fault-tolerant and highly parallel manner. Figure~\ref{fig:scene-reconstruction} shows \emph{RESFlow} tile inference and reconstruction illustration. The tile partitioning strategy injects a key property that allows for spatially non-contiguous image tiles to be processed by a single bucket model - enabling consistent inference over wide geographic conditions.

\subsection{Application space}
By design, the modules presented thus far within the \emph{RESFlow} framework have been agnostic to any specific application or use case within the realm of remote-sensing imagery. In this manner, one could think of the Model Gallery partitioned to fulfill multiple tasks such as object detection, neighborhood and settlement mapping, or temporal change detection, which are only a subset of potential applications. Based on this premise, within the Application Space multiple copies of the Model Gallery are formed, each containing trained models which are uniformly purposed to perform a given task.

\subsection{Image analytics via parallel computing}
\begin{figure*}
    \centering
    \includegraphics[width=\linewidth,height=.4\linewidth,]{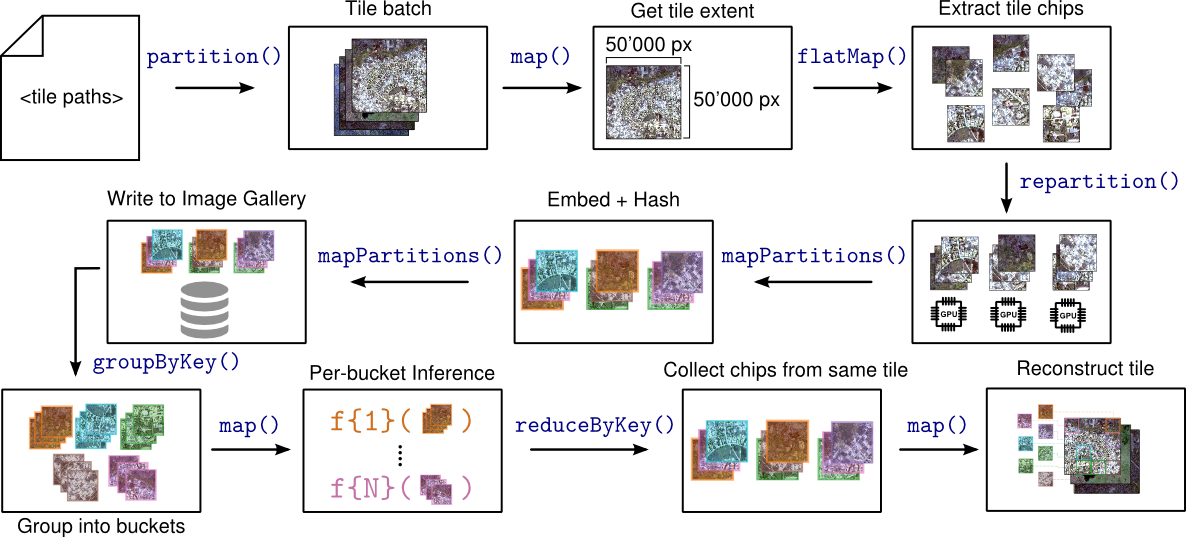}
    \caption{\emph{RESFlow} parallel inferencing flow: a high-level representation of the Spark based transformations and actions implementation.}
    \label{fig:spark-layout}
\end{figure*}%
\subsubsection{Satellite Imagery RDDs} 
Spark currently does not support extended image file types such as .tiff or .dicom to be serialized into byte arrays encapsulated within its RDD objects~\footnote{see: \url{https://issues.apache.org/jira/browse/SPARK-21866}}. As a consequence of this limitation, a design choice was required between either converting the collected sensor-based imagery used within \emph{RESFlow} into supported formats (such as 8-bit .jpg), or to instead use RDD objects to store path-based references to the location of the imagery stored within network storage. The former approach would result in a loss of data precision (32-bit to 8-bit for each scene), while the latter would force the image data to be read twice from disk during workflow execution (once for hashing and embedding, and a second time for per-bucket inference). With precedence being placed on precision to enable higher levels of accuracy during training, we choose the latter option and instantiate an RDD to contain the paths of the scene data to be analyzed in \emph{RESFlow}. 

\subsubsection{User Defined Functions} %
We implement several User Defined Functions~(UDFs) within Spark in order to realize \emph{RESFlow's} operation. Represented in Figure~\ref{fig:spark-layout}, these functions encapsulate tasks such as getting the extent of a given scene tile, or performing inference across a partition-based bucket. We utilise external libraries such as Tensorflow~\cite{abadi2016tensorflow}, Pytorch~\cite{paszke2017automatic} and GDAL~\cite{GDAL} to assist in performing these operations which act upon one or more rows records within a given RDD partition.   

\subsubsection{GPU Allocation Heuristic}\label{sec:gpu-checkout} 
An important limitation in using Spark for \emph{RESFlow}'s implementation is its lack of support for GPU-based resources. Here, first-class status is given to cluster-based resources such as CPU cores and RAM; allowing a fixed quantity of these resources to be allocated to instantiate a spark-executor. On the other hand, Spark contains no functionality to enable GPUs to be exclusively associated with a given executor in the same manner. When considered in the light of how Spark runs tasks which belong to the same processing stage, maintaining independence and concurrency, this lack of GPU resource allocation can become problematic. For example, during the embedding and hashing step of the workflow each executor is required to process the data within its associated partitions via GPU computation. With no robust mechanism to reserve a GPU for a given executor, this step may result in several executors utilizing the same GPU - causing the stage to fail as the limited amount of GPU RAM is quickly exhausted by the concurrent executor tasks.

As a workaround to this problem, we implemented a simplistic `GPU checkout' routine which allows for the soft-assignment of GPU resources to take place between the executors on a given worker node. We accomplish this by creating a ticketing folder within the executors' shared work-space, into which we place one or more `tickets' per physical GPU within the node. Using this ticket-system metaphor, when an executor requires the use of a GPU, it scans the ticketing folder for available instances - removing a ticket if one is available, or returning a ticket if the GPU resource is no longer required.   

\section{Experiments and Results}\label{sec:experiment-workloads}
\begin{figure*}
    \centering
    \includegraphics[width=.6\linewidth]{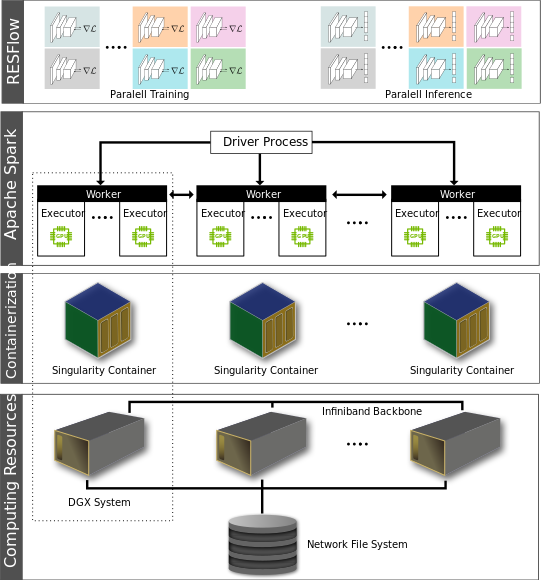}
    \caption{System configuration for \emph{RESFlow}.}
    \label{fig:hardware-stack}
    \vspace{-.1in}
\end{figure*}%

\emph{Systems specifications:}The computing environments for all experiments consist of various Spark cluster configurations on GPU optimized platforms. All Spark clusters are configured to take advantage of the NVIDIA DGX Systems as they deliver an integrated hardware and software solution that’s been optimized to deliver faster time-to-solution with latest GPU resources.
\emph{Nvidia-DGX1 systems:} We first consider an environment with 3xNvidia-DGX1 machines each capable of a total of  80 threads from 40 cores via two Xeon CPU E5-2698 processors operating at 2.20GHz with 512G DDR4 of system RAM. Each machine has eight 16GB Volta GPUs achieving a maximum graphics clock of 1530Mhz and a maximum memory clock speed of 877Mhz with a  300Watt power cap on each GPU.  The all SSD based local storage has 500G for the OS with a additional 7TB on Raid 0 for data storage. Each machine is connected to Network File System (NFS) storage via a  single 10Gb  Ethernet network connection. In total we have three such machines that are interconnected via 4x100Gb EDR IB ports (2 GPUs per IB connection).
\emph{Nvidia-DGX2 systems:} For the second environment we consider cluster nodes setup on NVIDIA DGX2 systems which each combine 16 GPUs fully interconnected via NVLink. The first node, a Nvidia-DGX2, can run  96 threads from 48 cores via two Xeon Platinum 8168 processors operating at 2.7Ghz with 1.5TB of DDR4 system RAM. The machine has sixteen 32GB Volta 350watt GPUs with max clocks of  1597/958 (Graphics/Memory). Local storage consist of a 1TB NVME Raid 1 boot partition and  a  30TB NVME  Raid0 data partitoin offering maximum  transfer speeds of approx 20GBps. The machine has 8x100Gb EDR IB Ports (2 GPUs per IB port). The second node, Nvidia-DGX2-H, similarly has 96 threads with 48 cores, however with two Xeon Platinum 8174 operating at 3.1Ghz with 1.5TB RAM DDR4 2666. The machine has sixteen 32GB OCed Volta GPUs running at 450watt maximum power consumption with max clocks of 1702/1107 (Graphics/Memory),  local storage consists of a 1TB NVME Raid 1 boot partition and  a 30TB NVME  Raid0 data partition offering maximum  transfer speeds of approx 20GBps. The machine equally has 8x100Gb EDR IB Ports (2 GPUs per IB port).

Experimental evaluations are conducted by first setting up a {\em single-Nvidia-DGX1 Spark cluster} as follows: a configuration of one master node and eight workers. The cluster is instantiated from a single Singularity containers as depicted in Figure~\ref{fig:hardware-stack}. 
Each worker is allocated 30GB memory and 9 cores and is responsible for launching a single executor. At execution the master node runs the driver process whose allocated memory is 10GB. Extending this configuration to a {\em three-node Nvidia-DGX1 Spark cluster} we instantiate a Singularity container with single master node and eight workers on one Nvidia-DGX1 machine and add two more Singularity instances on two more additional machines each equipped to support eight workers with allocation of 30GB memory and 9 cores per worker. Two more clusters, {\em single-node Nvidia-DGX2-H Spark cluster} and  {\em two-node Nvidia-DGX2 Spark cluster}, are setup in a similar manner however each worker is allocated 80GB of memory and 6 cores per worker.

\subsection{Case Study: Building Footprint Mapping}
Country scale building footprint mapping fits the scope of both a data and compute intensive application. We envision current deep learning algorithms for solving this case study benefiting from a hybrid combination of the in-memory computing capability of Apache Spark and high performance computing hardware platforms. 

Our experimental dataset consists of $(.3-.7)$-meter resolution satellite imagery acquired by Digital Globe constellations i.e. WorldView-2, and WorldView-3. These constellations provide high resolution imagery that is suitable for pixel level semantic segmentation objects. As summarized in Table~\ref{tab: Data-sources}, training and validation image data (using a 90\%:10\% split) is collected from several countries Ethiopia, South Sudan, Zambia. Each sample is a $500\times500~$ pixel RGB image with an associated label mask. Importantly, we further explore out-of-country/out-of-sample testing regions from other different geographical areas.These areas consist of New Mexico of the United States, Puerto Rico, Alabama, and Arizona, and are used to demonstrate both the deployment and generalization performance of {\em RESFlow}. These testing sets also represent a more realistic use-case of the framework once placed in production where rapid inferencing is much needed. Table~\ref{tab: Data-sources} summarizes the distribution of testing samples that are used from each of the out-of-sample locations. 
\begin{table}[]
\centering
\caption{Sources and number of labelled data for training, validation and testing evaluation.}
\label{tab: Data-sources}
\scriptsize
\begin{tabular}{@{}cccc@{}}
\toprule
Country/State & Data Split \\ \cmidrule(l){2-3} 
                                                        & Train samples          & Validation/Testing samples         \\ \midrule
Ethiopia                                       & 2714           & 302           \\
South Sudan                                      & 892            & 99            \\
Yemen                                            & 4023           & 447           \\
Zambia                                        & 1125           & 125           \\
Alabama                                          & 0              & 156           \\
Arizona                                            & 0              & 269           \\
Puerto Rico                                            & 0              & 300           \\
New Mexico                                       & 0              & 61\\ \hline
\end{tabular}
\end{table}

\subsection{Workflow initialization}
To initialize the {\em RESFlow} ensemble with all available training samples, we selected an optimal count of six buckets for the image and model galleries based-off the average intra-cluster variance measured over a range of cluster numbers used for the clustering and embedding module. For each of these buckets, we trained four different convolutional neural networks for the building mapping task where training and validation data are from its corresponding image gallery bucket. We picked these CNNs, namely~\textbf{ResNet50-FCN}, \cite{he2016deep}, \textbf{U-net}  \cite{ronneberger2015u}, \textbf{Seg-Net} \cite{7803544}, and \textbf{DeepLab} \cite{chen2018deeplab} without multiple feature fusion, based on various sizes of model (number of parameters to train) as well as their superior performance on semantic segmentation tasks. The number of parameters for these models are listed in Table \ref{tab: model-parameters-train}. This also showcases the flexibility of the  modularized application space in {\em RESFlow} where researchers can easily set up the preferred algorithms to test the model gallery. During testing (performing inferencing) on the out-of-sample data, each of these trained models becomes responsible for independently performing building extraction on testing samples,  which are assigned to its membership via the same process of image gallery mapping. When reporting results, we refer to this combined quorum of models as \emph{RESFlow}. All of the CNNs were concurrently trained from scratch, without using any pre-trained models. Hyper-parameters are held constant through all experiments so that we limit number of variables in the experimental runs.

\begin{table}[b]
\caption{Number of parameters and training time for used CNNs. Speed decreases with model complexity.}
\scriptsize
\begin{tabular}{cccc}
\hline
           & Model parameters & time per sample(sec) \\
           \hline
ResNet50-FCN & 23,541,769   & 0.032                          \\ 
U-Net      & 13,395,329       & 0.068                          \\ 
Seg-Net     & 29,443,073       & 0.076                           \\ 
DeepLab    & 42,542,280       & 0.111  \\   
\hline
\end{tabular}
\label{tab: model-parameters-train}
\end{table}%

\subsection{Performance and computational efficiency}
Among the main contributions of the paper, improving workload computational efficiency is vital in processing large volumes of data. RESFlow seeks to achieve this by combining algorithmic innovations as well as accelerated deep learning computing capable system architectures for remote sensing data analytics. More specific, its capability to partition data in the metric space offers desirable properties for large scale accelerated training and inferencing tasks. We observe and assess the system behavior under varying workloads and different computing environments to identify computational trade-offs between constrained resources and need to process large volumes of imagery. The deep learning fully convolutional network of choice is the \textbf{U-Net} architecture~\cite{ronneberger2015u}, merely selected for its fast training convergence across the $60\%$ F1-score on validation of $300$ samples as shown in Figure~\ref{fig:training-both-75epochs}.

To provide an appropriate comparison of performance we additionally train a single \textbf{U-Net} model, again using all available training samples with the identical fully convolutional network architecture. We refer to this monolithic network as the \emph{Mono} model, and measure its performance across the entire out-of-sample test set. As metrics, we utilize the Intersection over Union~(IoU) in addition to F1 scores in order to follow the accepted community standard~\cite{Demir2018} and report performance results in Table~\ref{tab: Held-out-test}. To quantitatively assert the viability of our proposed framework, Table~\ref{tab: Held-out-test} presents the performance of both the Mono and \emph{RESFlow} models on the held-out set of test data. Here \emph{RESFlow} is seen to perform very similarly to the Mono model for two of the three test regions. This is considerable, as each model from the \emph{RESFlow} Image Gallery sees considerably less data compared to its Mono model counterpart during training, and yet is able to generalize to a similar degree. This is not true for the New Mexico region, however, with a large deficit in performance being observed. Furthermore, visual maps to illustrate the performance of \emph{RESFlow} ensemble of models over large geographic extents is shown in Figure ~\ref{fig:building-mapping-large scale}.

\begin{figure*}
\centering
\includegraphics[width=\linewidth]{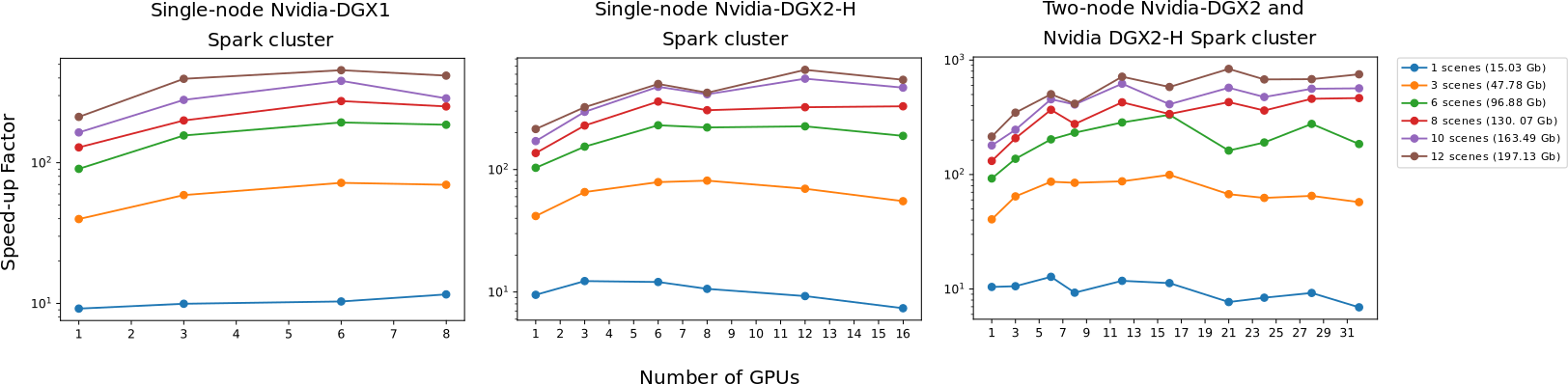}
\caption{A log-scale speed up of varying workloads when processed with {\em RESflow} parallel inference. Using serial processing single GPU baseline that averages inference speed of $35$minutes per $40'000\times35'000$ pixel image scene.}
\label{fig:speed-up-plot}
\end{figure*}

Table~\ref{tab:one-node-cluster} and Table~\ref{tab:two-node-cluster} illustrate the computational performance of deploying RESFlow on an Apache Spark cluster equipped with deep learning modules based on PyTorch and Tensorflow. The baseline is recorded to average 35 minutes for pixel-level semantic segmentation inferencing time for a single image scene of size 40,000x35,000 pixels and data volume of 11GB processed on a single Nvidia DGX1 V100 GPU. Figure~\ref{fig:speed-up-plot} shows the speedup as a function of both the number of GPUs and data sizes in GigaBytes(computed from number of image scenes). The speedup is calculated as the ratio of the baseline execution time to the normalized compute time for given number of GPUs.  For all different size workloads, we observe a tremendous speed up ranging from $9x$ to over $400x$ across all Spark cluster environments. Overall, Table~\ref{tab:one-node-cluster} to Table~\ref{tab:one-node-cluster-dgx2} and Figure~\ref{fig:speed-up-plot} demonstrate the speedup factors of the various cluster configurations. 

While we report on the strong and weak scaling of the segmentation inference module. The embedding and hashing modules are embarrassingly parallel with operations performed at image patch level to completion and always benefiting from additional numbers of GPU workers. To establish the strong scaling aspect of the workflow, Figure~\ref{fig:speed-up-plot} shows speed-up performance over 6 workloads of data (between 1-12 image scenes) while varying the number of spark-based workers. The inference module is parallelizable while the merging of image patches to reconstruct the large scene is performed in a serial fashion; perhaps introducing an upper limit on the compute speed-up. An immediate observation is that compute speedup does not assume a linear increase with additional GPUs - parallelization efficiency decreases for each workload as the amount of GPU workers are increased. The presence of the increasing and decreasing trends on the performance curves could be explained by highlighting two aspects. The first is due to algorithmic implementation of our workflow. As aforementioned the most important contribution of our workflow is its ability to partition or tile large imagery and enable computing at scale. During inference execution on tiled partitions, worker processors do not communicate with one another, however on completion, we perform a \textit{reduceByKey} operation within Spark to group all tiles of the same image scene ID for merging into the large extent output mask. The merging operation for output mask reconstruction is performed by a single worker. As illustrated from Figure~\ref{fig:speed-up-plot}, the performance bottleneck is more pronounced for workers executing on 6 GPUs or more. In addition, inferencing of fewer image scene appears to incur the largest compromise on speed up. The second aspect significantly influencing the observed performance bottleneck is attributed to increased I/O read activities. The computational efficiency introduced by RESFlow benefits tremendously from the lazy evaluation of Spark RDD data transformation. Throughout the pipeline, image data is only read from disk at inference time for each corresponding image tile. As a result, an increase in workers executing on more GPUs concurrently, thereby adding to the I/O read count,  considerably compounds and causes a decrease in compute performance. Table~\ref{tab:one-node-cluster} to Table~\ref{tab:two-node-cluster} further illustrate computational efficiency across a range imagery data sizes (or as measured in land area square kilometers). To establish weak scaling, we increase the workload on each GPU processor and also observe an improved weak scaling aspect of the workflow. Each row shows the compute time obtained for different area sizes for a fixed number of GPUs. The results indicate a desirable scaling factor, i.e. computational efficiency appears to increase as the land area to map (calculated for number of image scenes) correspondingly increases. For example, in Figure~\ref{fig:speed-up-plot}, for 12 GPU-workers, the compute speed for an area of $313.97$ sq.km (or a single image coverage with data size of $15.03$GB) is a $6.91x$ improvement over the baseline of 35minutes to process a single image scene. However, for an increased workload or land area of $4141.16$ sq.km(or 12 image scenes with data size of $197.13$GB), the speed-up reaches $750x$ over the baseline.
This performance increase appears to be due to fact that overall communication and system bottlenecks are more dependent on the number of GPUs than on the land area.

\begin{figure}
\centering
    \includegraphics[width=0.45\textwidth]{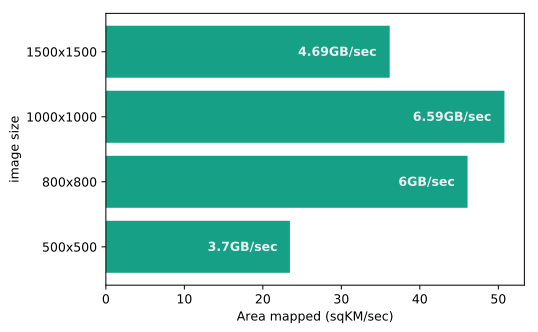}
    \caption{Illustration of total area mapped per second for different image input sizes on a Nvidia-DGX2-H Spark cluster.}
    \label{fig:DGX2-H-area-mapped}
\end{figure}

We also evaluate the overall inference performance for varying input image size to the \textbf{U-Net} network architecture. Figure~\ref{fig:DGX2-H-area-mapped} provides results for RGB image input of size $500\times500\times3$, $800\times800\times3$,$1000\times1000\times3$, and $1500\times1500\times3$. Inference is done in batches of size 12, 8, 5, and 2 image scenes, respectively. Compute efficiency is observed to vary  with network completing inference at rate $1220$ per sec (or throughput of $3.7GB/sec)$ for image tiles of size $500\times500\times3$, which has an equivalent area rate of $23.45$sq.km/sec. Given the smaller size in input image,  I/O reads are increased putting a burden to the NFS file system. By considering much larger input images not only do we reduce the amount of I/O reads per image scene but we also increase both the throughput and the equivalent land area mapped. For example, with $1000\times1000\times3$ input images the network reaches a peak inference throughput of $6.59GB/sec$ (processing a total of $564$ images/sec) and equivalence of mapping $50.78$sq.km/sec. At this rate we posit that the GPU compute time dominates the I/O read. However when for much larger input image sizes  we note a throughput degradation to level of $4.69GB/sec$ (processing a total of $178$ images/sec) for $1500\times1500\times3$ tiles. Even though larger tiles increase the GPU utilization time they also require allocation of larger memory footprint for the inference results.%

\begin{table*}[]
\centering
\caption{Compute efficiency(in mins) on varying workloads on One-node Nvidia-DGX1 Spark cluster.}
\label{tab:one-node-cluster}
\begin{threeparttable}
\scriptsize
\begin{tabular}{lllllllllr}
\hline
&&\multicolumn{4}{c}{Files size(GB)} \\
\cline{2-7}
   & 15.03 & 47.78 & 96.88 & 130.07 & 163.49 & 197.13 \\
  \cline{2-7}
&&\multicolumn{4}{c}{Area (sq.km)} \\
\cline{2-7}
Workers    & 313.97  & 1002.15 & 2035.66 & 2733.05 &3433.78  & 4141.16 \\
\hline
1GPU      & 3.81m & 7.92m & 13.93m & 17.43m\tnote{*} & 21.32m\tnote{*} & 23.81m\tnote{*}\\
3GPUs    & 3.51m & 5.35m & 8.07m & 11.22m & 12.52m & 12.77m \\
6GPUs    & 3.39m & 4.38m & 6.52m\tnote{*} & 8.17m\tnote{*} & 9.17m\tnote{*} & 11.09m\tnote{*} \\
8GPUs   & 3.02m\tnote{*} & 4.52m\tnote{*} & 6.77m\tnote{*} & 8.92m\tnote{*} & 12.21m\tnote{*} & 12.12m\tnote{*}\\
\hline
\end{tabular}
\begin{tablenotes}
\item[*] Indicates a result for which soft gpu-executor assignment is needed to prevent run failure. 
\end{tablenotes}
\end{threeparttable}
\end{table*}%

\begin{table}[]
\centering
\caption{Results for held-out testing data.}
\label{tab: Held-out-test}
\scriptsize
\begin{tabular}{clcc}
\toprule 
\textbf{Region} & \textbf{Model} & \textbf{IoU} & \textbf{F1-score}\\
\hline
Alabama & Mono & 0.64 & 0.78\\
 & RESFlow & 0.63 & 0.77\\
\hline
Arizona & Mono & 0.79 & 0.88\\
 & RESFlow & 0.76 & 0.86\\
\hline
Puerto Rico & Mono & 0.54 & 0.70\\
 & RESFlow & 0.52 & 0.69\\
\hline
New Mexico & Mono & 0.72 & 0.84\\
 & RESFlow & 0.62 & 0.77 \\
 \hline
\end{tabular}
\end{table}%
\begin{table*}[]
\centering
\caption{ Compute efficiency(in mins) on varying workloads on One-node Nvidia-DGX2-H Spark cluster.}
\label{tab:one-node-cluster-dgx2}
\begin{threeparttable}
\scriptsize
\begin{tabular}{lllllllll}
\hline
&&\multicolumn{4}{c}{Files size(GB)} \\
\cline{2-7}
   & 15.03 & 47.78 & 96.88 & 130.07 & 163.49 & 197.13 \\
\cline{2-7}
&&\multicolumn{4}{c}{Area (sq.km)} \\
\cline{2-7}
Workers    & 313.97  & 1002.15 & 2035.66 & 2733.05 &3433.78  & 4141.16 \\
\hline
1GPU      & 3.69m & 7.55m & 12.19m & 16.43m & 20.50m & 23.53m \\
3GPUs     & 2.87m & 4.81m & 8.19m & 9.79m & 11.82m & 15.6m\\
6GPUs    & 2.9m & 3.99m & 5.48m & 6.24m & 7.38m& 10.08m\\
8GPUs    & 3.3m & 3.88m & 5.71m& 7.34m& 8.5m & 11.89m\\
12GPUs   & 3.78m & 4.52m & 5.59m\tnote{*} & 6.94m\tnote{*}& 6.34m\tnote{*} & 7.73m\tnote{*}\\
16GPUs  & 4.75m\tnote{*}  & 5.71m\tnote{*} & 6.69m\tnote{*} & 6.82m\tnote{*} & 7.50m\tnote{*} & 9.32m\tnote{*}\\
\hline
\end{tabular}
\begin{tablenotes}
\item[*] Indicates a result for which soft gpu-executor assignment is needed to prevent run failure.
\end{tablenotes}
\end{threeparttable}
\end{table*}%

\begin{table*}[]
\centering
\caption{Compute efficiency (in mins) on varying workloads on Two-node Nvidia-DGX2-H Spark cluster.}
\label{tab:two-node-cluster}
\scriptsize
\begin{tabular}{lllllllll}
\hline
&&\multicolumn{4}{c}{Files size(GB)} \\
\cline{2-7}
   & 15.03 & 47.78 & 96.88 & 130.07 & 163.49 & 197.13 \\
\cline{2-7}
&&\multicolumn{4}{c}{Area (sq.km)} \\
\cline{2-7}
Workers    & 313.97  & 1002.15 & 2035.66 & 2733.05 &3433.78  & 4141.16 \\
\hline
1GPU    & 3.36m & 7.76m & 13.64m & 17.04m & 19.49m & 23.49m \\
3GPUs   & 3.31m & 4.89m & 9.18m & 10.79m & 14.22m & 14.51m\\
6GPUs   & 2.74m & 3.64m & 6.21m & 6.09m & 7.7m & 10.01m\\
8GPUs   & 3.77m & 3.72m & 5.43m & 8.11m & 8.55m & 12.14m\\
12GPUs  & 2.97m & 3.61m & 4.42m & 5.24m & 5.63m & 7.04m\\
16GPUs  & 3.11m & 3.17m & 3.79m & 6.63m & 8.51m & 8.66m\\
21GPUs  & 4.55m & 4.68m & 7.78m & 5.22m & 6.11m & 6.02m\\
24GPUs  & 4.17m & 5.04m & 6.61m & 6.18m & 7.36m & 7.44m\\
28GPUs  & 3.79m & 4.84m & 4.55m & 4.88m & 6.23m & 7.40m\\
32GPUs  & 5.06min & 5.49m & 6.82m & 4.81m & 6.18m & 6.71m\\
\hline
\end{tabular}
\end{table*}

\section{Large scale experiments}\label{sec:large-experiments}
\begin{figure}    
    \includegraphics[width=\linewidth]{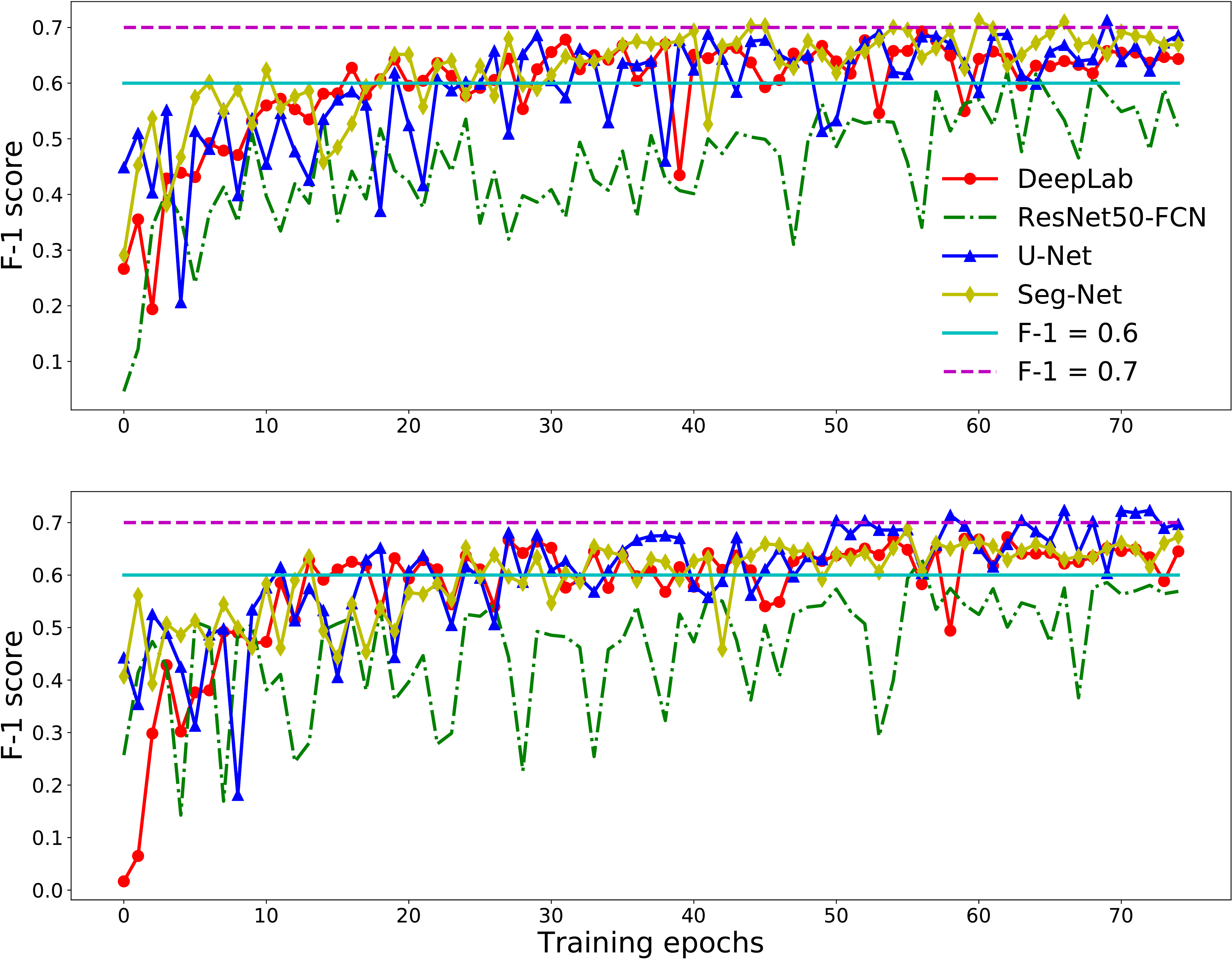}
    \caption{F-1 scores on validation set for the tested CNNs. The models were trained on a regular DGX-1 machine (top) and on a SPARK GPU cluster (bottom).}
\label{fig:training-both-75epochs}
\end{figure}

\begin{figure*}
    \centering
    \includegraphics[width=1\linewidth]{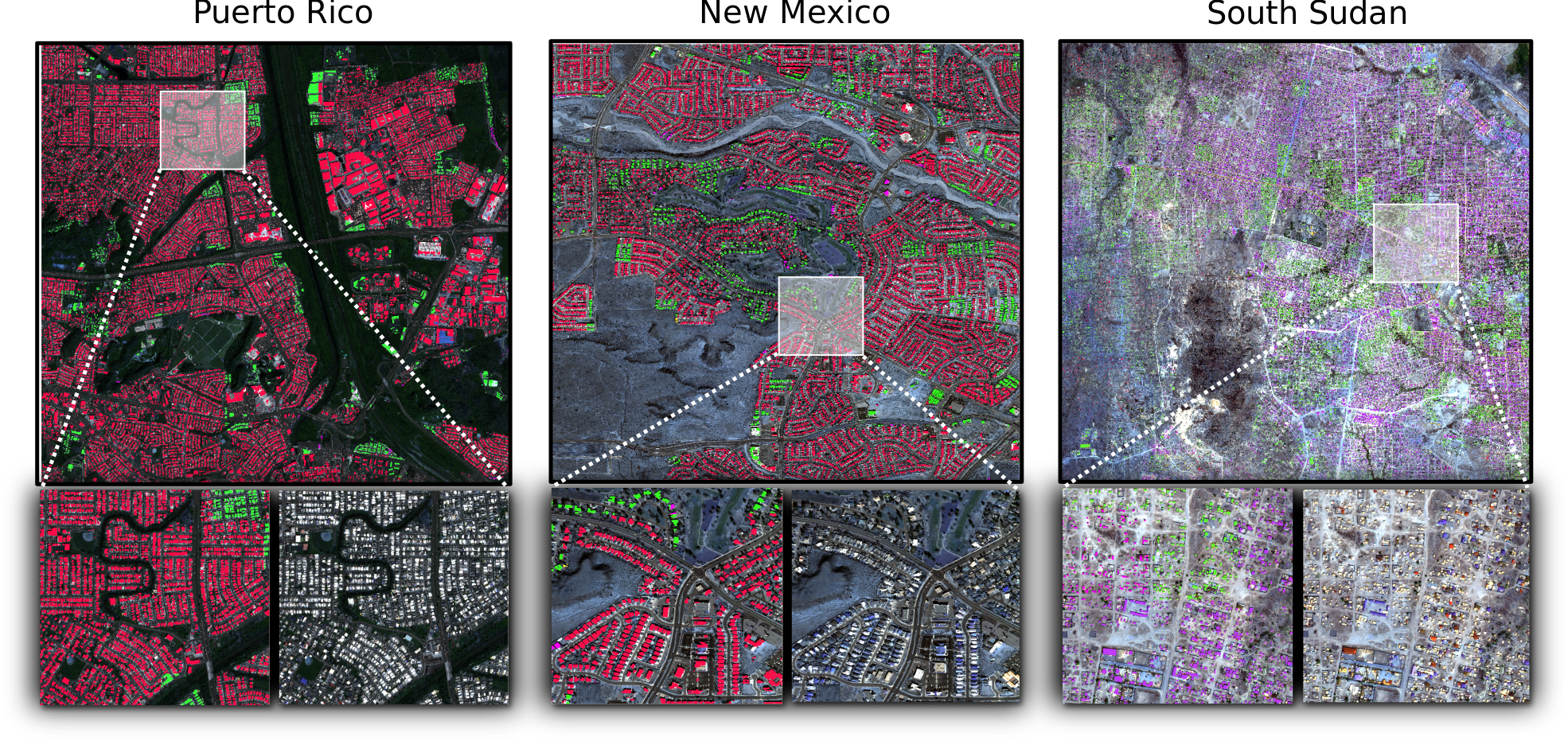}
    \caption{Example building footprint mapping results for the \emph{RESFlow} models on held-out New Mexico, Puerto Rico and South Sudan data. Owing to the varying image characteristics and geographies image tiles clipped from same image scene are processed in parallel using different bucket models. \tiny{Note: varying mask colours denote different bucket models for a total of six models from the model gallery.}}
    \label{fig:building-mapping-large-scale}
\end{figure*}

\paragraph{Training:}We first evaluate several aspects of the training phase in the SPARK-enabled GPU clusters. Taking advantage of the partitioned image gallery, there is no interaction between buckets, turning the model training into an embarrassingly parallel task. As shown in Table \ref{tab: model-parameters-train}, the training speed scales with the complexity of the model (i.e. number of trainable parameters). Note that also the total time needed to achieve the model convergence is also linear to the number of training samples ($\text{total training time} = \text{number of training samples} \times \text{secs/ sample})$). In Figure \ref{fig:training-both-75epochs}, we demonstrate the similar model training performance achieved by a regular DGX-1 and a SPARK-enabled DGX-1. Because of different network initialization conditions (as we trained the models from scratch), the learning curves are not identical under these two computing environments. Nevertheless, with the fixed training hyper-parameters, we can see the best F-1 scores obtained from both computing platforms for these four models are similar: \textbf{Seg-Net} and \textbf{U-Net} both deliver F-1 scores slightly above 0.7, \textbf{DeepLab} has F-1 scores close to 0.7, and \textbf{ResNet50-FCN} achieves F-1 scores close to 0.65. 
In addition, the training improvements become trivial after $\sim$25 epochs for both computing environments except for \textbf{ResNet50-FCN}, which requires longer training epochs ($\sim$60 epochs) to get optimized parameters.
\paragraph{Inference:} We here present the large scale inferencing results produced by the {\em RESFlow} model gallery on three states/countries: Puerto Rico, New Mexico of the United States and South Sudan. Figure~\ref{fig:building-mapping-large-scale} visualizes output building semantic segmentation maps across varying geographies that encompass the above three test sites. Finally, having evaluated the different components for \emph{RESFlow} e.g. impact of varying number of workers for different number of image to process in a single batch (Figure~\ref{fig:speed-up-plot}), establishing the throughput bounds as a function of input image size (Figure~\ref{fig:DGX2-H-area-mapped}), we assess the scalability and applicability of deploying the workflow as depicted in Figure~\ref{sec:sequential-workflow} on $\mathbf{14}$TB of imagery data covering the State of New Mexico. We select the Two-node Nvidia-DGX2 Spark cluster to execute the task with $28$GPUs and batch size of $12$ for image scenes.  The pipeline execution entails computing three deep learning tasks for each image tile: deep feature extraction stage, a deep hashing stage and a deep semantic segmentation inference. The main goal for the large scale deployment was to assess the performance of our pipeline when deployed for production task. Therefore we account for both read and write (I/O) bottlenecks in addition to the compute time. Table~\ref{tab:production} presents the throughput for this workload.
\begin{table}[hb!]
\centering
\caption{\emph{RESFlow} deployment performance on Two-node Nvidia-DGX2 Spark cluster.}
\label{tab:production}
\scriptsize
\begin{tabular}{clcc}
\toprule 
 & \textbf{Per second} & \textbf{Per day} \tabularnewline
\midrule
Area mapped (sq.km) & 5.245 & ~453,168 \tabularnewline
\midrule 
Number of images & 80 & ~6,912,000 \tabularnewline
\midrule 
Total image data (GB) & 0.243 & ~21,028\tabularnewline
\hline
\end{tabular}
\end{table}%

This end-to-end inferencing workflow demonstrates unprecedented processing of vast amounts of satellite imagery. Averaging a throughput of $.243GB$/sec (or $5.245$sq.km/sec) inference of the entire set of $1440$ image scene completed in $21$hrs - a tremendous achievement over a previous serial based inference workflow that would have taken over 28days to complete.%

\section{Conclusions and Future Work}\label{sec:conclusions}
With a novel remote sensing data partitioning concept, this paper presented a parallel inferencing workflow based on accelerated AI deployment hardware. Demonstrated on the pixel labeling challenge, we extended herein, Apache Spark based satellite imagery RDDs for processing with deep learning at  scale and demonstrated compute efficiency across Terrabytes of high resolution data covering land area equivalent of $787'300$ sq.km. We achieve unprecedented pixel labeling area rates of 5.245sq.km/sec, amounting to 453,168 sq.km/day (or a daily capacity processing of 21,028Terrabytes) demonstrating a reduction of a 28 day workload to 21hours. In order to leverage Apache Spark to support deep feature learning and accommodate the problem of model generalization, the remote sensing data partitioning, the central concept in {\em RESFlow}, is realized from a combined set of four modules. {\em RESFlow} ignites optimism about a number of other relevant, and perhaps such new research directions, including enabling faster search for retrieval of images with similar content, combining human geography and machine learning e.g. establishing limiting bounds and performance of models learned from finer abstract and deconflated geographical spaces, establishing robustness and stability of learning models for objects that are rare in some geographies and common in others. These directions involve studying varying levels of imagery data complexity and their impact on training and inference tasks, and can potentially benefit from sensor and geographic agnostic workflows as compared to spatially constrained approaches.
\emph{Other future directions:} understanding the staging of geospatial workloads and deploying containerized workflows on supercomputing platforms presents a future research direction for our work. In addition, the work can be extended by exploring methodologies to enable better architectural design of computers specific to geospatial processing to benefit more applications.

\section*{Acknowledgement}
Additionally, we would like to acknowledge that this manuscript has been authored by UT-Battelle, LLC under Contract No. DE-AC05-00OR22725 with the U.S. Department of Energy. The United States Government retains and the publisher, by accepting the article for publication, acknowledges that the United States Government retains a non-exclusive, paid-up, irrevocable, world-wide license to publish or reproduce the published form of this manuscript, or allow others to do so, for United States Government purposes.
%

\bibliographystyle{IEEEbib}
\bibliography{resflow}

\end{document}